\documentclass[sigconf]{acmart}
\AtBeginDocument{%
  }

\setcopyright{acmlicensed}
\copyrightyear{2025}
\acmYear{2025}
\acmDOI{XXXXXXX.XXXXXXX}
\acmConference[KDD-UMC '25]{The Undergraduate and Master’s Consortium at KDD 2025}{August 03--07,
  2025}{Toronto, Canada}
\acmISBN{978-1-4503-XXXX-X/2018/06}

\usepackage{algorithm}
\usepackage{algpseudocode}
\usepackage{amsthm,amsmath}
\usepackage{graphicx}     
\usepackage{tikz}   
\usepackage{subcaption}    
\usepackage{caption}      
\usetikzlibrary{arrows.meta,calc}
\newcommand{\features}{\mathcal{F}}
\newcommand{\feature}{f}
\newcommand{\aug}{\mathcal{A}}

\newcommand{\user}{\mathcal{U}}

\newcommand{\items}{\mathcal{I}}
\newcommand{\rules}{\mathcal{R}}
\newcommand{\treeRoot}{\text{root}}

\newcommand{\model}{C}
\newcommand{\capacity}{m}
\newcommand{\weights}{w}

\newcommand{\versionSpace}{\mathcal{V}}

\newcommand{\queries}{\mathcal{Q}}

\newcommand{\arule}{r}
\newcommand{\ball}{\mathcal{B}}
\newcommand{\heap}{\mathcal{H}}
\newtheorem*{theorem*}{Theorem}

\newcommand{\dist}{d}
\newcommand{\radius}{r}
\newcommand{\centerModel}{w_c}
\newcommand{\centerBall}{c}
\newcommand{\LB}{\text{LB}}
\newcommand{\UB}{\text{UB}}
\newcommand{\SDB}{\mathcal{D}}	

\newcommand{\CR}{\textsc{ChoquetRank }}
\begin{document}

\title{Geometry-Aware Active Learning of Pattern Rankings via Choquet-Based Aggregation}

\author{Tudor Matei Opran}
\email{tudor-matei.opran@imt-atlantique.net}
\orcid{1234-5678-9012}
\affiliation{%
  \institution{IMT Atlantique, LS2N, UMR CNRS 6004, F-44307 Nantes}
  \city{}
  \state{}
  \country{France}}

\author{Samir Loudni}
\email{samir.loudni@imt-atlantique.fr}
\orcid{1234-5678-9012}
\affiliation{%
  \institution{IMT Atlantique, LS2N, UMR CNRS 6004, F-44307 Nantes}
  \city{}
  \state{}
  \country{France}}

\renewcommand{\shortauthors}{Opran et al.}

\begin{abstract}
    We address the pattern explosion problem in pattern mining by proposing an interactive learning framework that combines nonlinear utility aggregation with geometry-aware query selection. Our method models user preferences through a Choquet integral over multiple interestingness measures and exploits the geometric structure of the version space to guide the selection of informative comparisons. A branch-and-bound strategy with tight distance bounds enables efficient identification of queries near the decision boundary. Experiments on UCI datasets show that our approach outperforms existing methods such as ChoquetRank, achieving better ranking accuracy with fewer user interactions.

\end{abstract}




\maketitle

\section{Introduction}

A central challenge in pattern mining is the pattern explosion problem~\citep{Agrawal94}: the number of patterns often grows exponentially, overwhelming the user with redundant results. To mitigate this, Interactive Learning of Pattern Rankings (ILPR)~\citep{LETOR-book,HumanInTheLoop} integrates expert feedback into the discovery process by learning a utility function from preference feedback rather than relying on fixed objective criteria.
In this work, we propose a two-stage interactive framework: in the first stage, a diverse set of association rules is extracted without ranking; in the second stage, we learn a user-specific preference model by aggregating multiple rule-interestingness measures using the Choquet integral~\citep{Marichal00,Grabisch:Choquet00}. This flexible nonlinear aggregation captures complex interactions between criteria, allowing nuanced modeling of subjective interest.
Our method distinguishes itself by treating learning as a geometric problem: every user comparison defines a separating hyperplane through the version space of candidate linear utility models. We guide query selection by identifying a representative center in this version space and focusing on rule pairs whose associated hyperplanes pass close to it—selecting queries of high uncertainty where user feedback is most valuable. To efficiently identify such comparisons, we introduce a branch-and-bound algorithm with tight geometric bounds on the distance between rule pairs and the central direction vector, enabling effective pruning and fast identification of informative queries.
We evaluate our approach against \CR~\citep{ChoquetRank}, a recent method that learns Choquet-based utilities from complete rankings. Experiments on several UCI benchmark datasets show that our method achieves more accurate rule rankings with fewer queries, demonstrating the benefits of combining nonlinear utility aggregation with geometry-driven active learning.

\section{Related Works}
Interactive pattern–sampling methods such as \textsc{LetSIP}~\citep{letsip}, \textsc{DISPALE}~\citep{HienLAOZ23} and \textsc{PRIIME}~\citep{PRIIME} rely on heuristic query–selection policies; they forgo active–learning guarantees, often citing the non-negligible cost of sampling~\citep{AIDE} or the NP-hardness of optimal query selection~\citep{Alon2006RankingT,Dzyuba,ChoquetRank}.  
By contrast, the active–learning literature offers provably sample-efficient algorithms for pairwise-comparison ranking, including noise-tolerant schemes~\citep{DBLP:journals/jmlr/Ailon12,NoiseTolerantPairwise,HerinNoiseTolerant}.  
A key algorithm in the noise-free \emph{realizable} setting is \emph{Generalised Binary Search} (GBS): first proposed for arbitrary hypothesis classes by Nowak~\citep{Nowak2008GeneralizedBS}, GBS greedily queries the sample that most evenly bisects the set of valid models, attaining optimal label complexity for linear separators under certain geometric conditions~\citep{mussmann2018}. Most recently, Di~Palma \emph{et al.}~\citep{DiPalma24} combined explicit version-space maintenance with randomised query sampling, demonstrating state-of-the-art label efficiency in human-in-the-loop model building in the \emph{realizable} setting.  Yet, to date, no interactive pattern-mining framework fully exploits GBS; we therefore build upon Choquet-based aggregation~\citep{ChoquetRank,opran2025echantillonnage} and integrate GBS-style querying to close this gap. 


\section{Preliminaries}
\noindent
\textbf{A) Association Rules and Choquet Integral}
Let \(\items\) be a finite set of items, an \emph{itemset} (or pattern) $X$ is a non-empty subset of $\items$. 
A transactional dataset $\SDB$ is a multiset of transactions over $\items$, where each \emph{transaction} $t \subseteq \items$. 
A pattern $X$ \emph{occurs} in a transaction $t$, iff $X \subseteq t$. 
The {\it frequency} of $X$ in $\SDB$ is the numbers of transactions that contain $X$.  
Patterns are fundamental in pattern mining and capture local structures (combination of items that co-occur frequently) or associations present in the data. An association rule is an implication  \(X\!\Rightarrow\!Y\) with \(X,Y\subseteq\items,\;X\cap Y=\varnothing\) asserting that transactions containing \(X\) tend also to contain \(Y\). 
Let \(\rules\) be a collection of candidate \emph{association rules}. Each rule is typically evaluated by objective interestingness measures such as support, the empirical frequency of \(X\cup Y\), and confidence, the conditional probability of \(Y\) given \(X\). More generally, we define a mapping \(\Phi:\rules\to\features\subset\mathbb{R}^{d}\), 
that associates each rule \(\arule \in \rules\) with a vector $\feature_\arule=(\feature_1,…,\feature_d)$ of $d$ such interestingness measures~\cite{tan_selecting_2004}.
We model the utility of a rule via a $k$-additive Choquet integral in Möbius form~\citep{Grabisch04}:

\[ \small
  \model_\capacity(\feature_r)
  \;=\;
  \sum_{\substack{A\subseteq[d]\\ 1\le |A|\le k}}
        \capacity(A)\,\min_{i\in A}\feature_{i},
\]
where the Möbius coefficients $\capacity:2^{[d]}\to\mathbb{R}$
encode all interactions of order at most \(k\) between criteria. Coefficients with \(|A|>k\) being set to zero. 
The Choquet integral generalizes weighted averages by capturing redundancy or synergy between features.
Monotonicity and normalisation of the underlying capacity impose the following \emph{linear} constraints on \(\capacity\)~\citep{Grabisch97,Grabisch2016}: 
\vspace*{-.1cm}
\begin{equation} \small
\label{eq:capacity-constraints}
\begin{cases}
\displaystyle
\sum_{\substack{T\subseteq S \\ |T|\le k-1}}
      \capacity\!\bigl(T\cup\{i\}\bigr) \;\ge 0,
& \forall\,i\in[d],\;\forall\,S\subseteq[d]\!\setminus\!\{i\},\\[8pt]
\displaystyle
\sum_{\substack{A\subseteq[d] \\ 1\le |A|\le k}}
      \capacity(A) \;=\; 1.
\end{cases}
\end{equation}

\noindent
\textbf{B) Interactive Learning of Pattern Rankings} 
In many data mining tasks, 
the number of candidate patterns or rules can be extremely large. However, only a small subset of these patterns are typically relevant to a particular user. Since relevance is inherently subjective and domain-dependent, it is often impractical to define an objective utility function a priori.
Interactive learning of pattern rankings addresses this challenge by involving the user in the loop~\citep{DzyubaLNR14,Dzyuba:letsip}. Rather than assuming fixed interestingness criteria, we aim to learn a utility function from user feedback that ranks patterns according to their subjective preferences~\citep{LETOR-book}.
Let \(\user: \rules \times\rules \mapsto \{-1,1\}\) be a binary preference oracle that expresses pairwise user preferences. The user is presented with pairs of patterns \(\arule_i, \arule_j \in \rules\) and provides comparative feedback:
$\user(\arule_i,\arule_j)=1$ if $\arule_i$ is preferred to $\arule_j$, \(-1\) otherwise. 
The goal is to learn an aggregation function \(\model_\capacity\) that aligns with the user’s preferences, such that:
\(
\forall \arule_i, \arule_j \in \rules, \quad \user(\arule_i, \arule_j) = 1 \Leftrightarrow \model_\capacity(\Phi(\arule_i)) \geq \model_\capacity(\Phi(\arule_j))
\). 
%
Each user preference induces the following constraint on the capacity $\capacity$ of the Choquet integral:
\vspace*{-.1cm}
\begin{equation} \small
    \label{eq:pref_constraints}
    \user(\arule_i, \arule_j) \times (\model_\capacity(\Phi(\arule_i)) - \model_\capacity(\Phi(\arule_j))) > 0
\end{equation}
ensuring that the aggregation function ranks the patterns consistently with the user’s expressed preference.
The set of all capacities that satisfy the accumulated constraints defines the \emph{version space} $\versionSpace$, i.e., the set of all valid\footnote{That is, capacities that induce the same ranking relations as the user feedback.} capacities.
To efficiently explore this space, the learning algorithm (Alg.~\ref{alg:gbs-choquet}) iteratively: 
(1) selects the next most informative rule pair $(\arule_i,\arule_j)$ to query, based on how evenly the current version space is split by their comparison. This selection strategy follows a greedy binary search (GBS)~\citep{DiPalma24} heuristic; (2) refines the version space $\versionSpace$ with the new constraint; (3) updates the model accordingly. 

Note that selecting an optimally informative query from a finite candidate pool \(\rules\times\rules\) at each step is equivalent to building an optimal binary decision tree, which is known to be NP-complete~\cite{NP-Complete-Trees}.


\begin{algorithm}[H] \small
\footnotesize
\caption{GBS for $k$-additive Choquet aggregation}
\label{alg:gbs-choquet}
\begin{algorithmic}[1]
\State \textbf{Input:} Candidate rule set $\rules$, user oracle $\user$
\State $\versionSpace \gets$ monotonic and normalized \(k\)-additive capacities
\For{$N$ learning iterations}
  \State $(r_i,r_j)\gets\textsc{SelectQuery}(\versionSpace, \rules)$
  \State $y\gets\user\!\bigl(\Phi(r_i),\Phi(r_j)\bigr)$
  \If{$y=1$}
      \State $\versionSpace\gets \versionSpace\cap\{m:C_m(\Phi(r_i))\ge C_m(\Phi(r_j))\}$
  \Else
      \State $\versionSpace\gets \versionSpace\cap\{m:C_m(\Phi(r_j))\ge C_m(\Phi(r_i))\}$
  \EndIf
\EndFor
\State \Return Aggregation model $\model_{\capacity^{*}} \leftarrow$ \(\textsc{center}(\versionSpace)\)
\end{algorithmic}
\end{algorithm}




\section{Geometrical Capacity Approximation}
\begin{figure*}[t] 
  \centering
  \begin{subfigure}[t]{0.24\textwidth}
    \includegraphics[width=\linewidth]{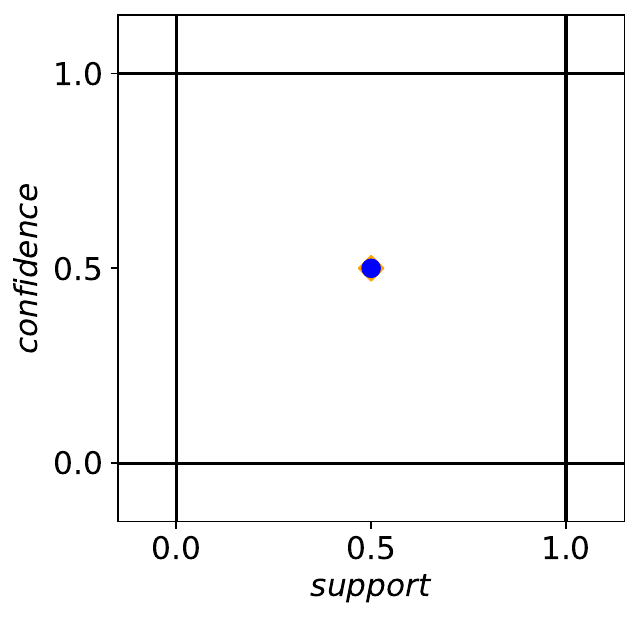}
    \caption{Initial constraints}
    \label{fig:init_cons}
  \end{subfigure}\hfill
  \begin{subfigure}[t]{0.24\textwidth}
    \includegraphics[width=\linewidth]{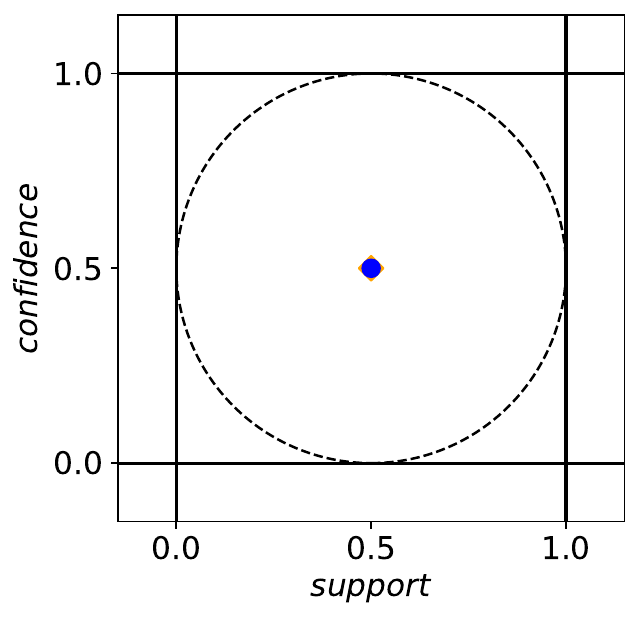}
    \caption{Maximal enclosed ball}
    \label{fig:max_ball}
  \end{subfigure}\hfill
  \begin{subfigure}[t]{0.24\textwidth}
    \includegraphics[width=\linewidth]{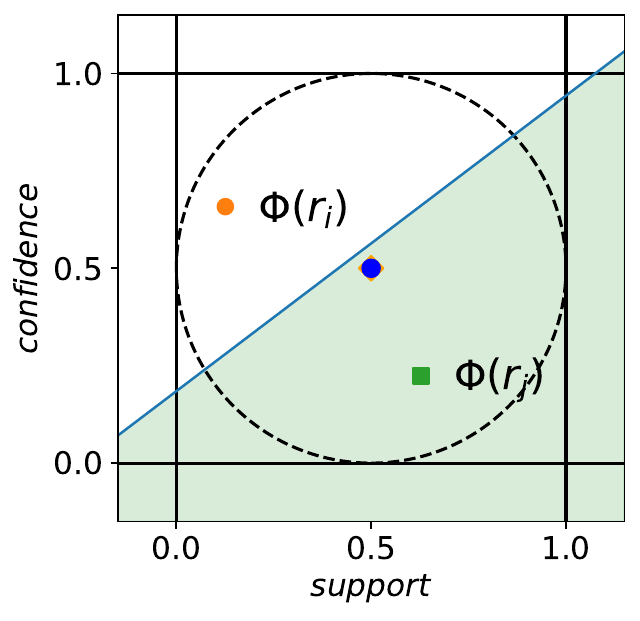}
    \caption{Preference constraint}
    \label{fig:pref_cons}
  \end{subfigure}\hfill
  \begin{subfigure}[t]{0.24\textwidth}
    \includegraphics[width=\linewidth]{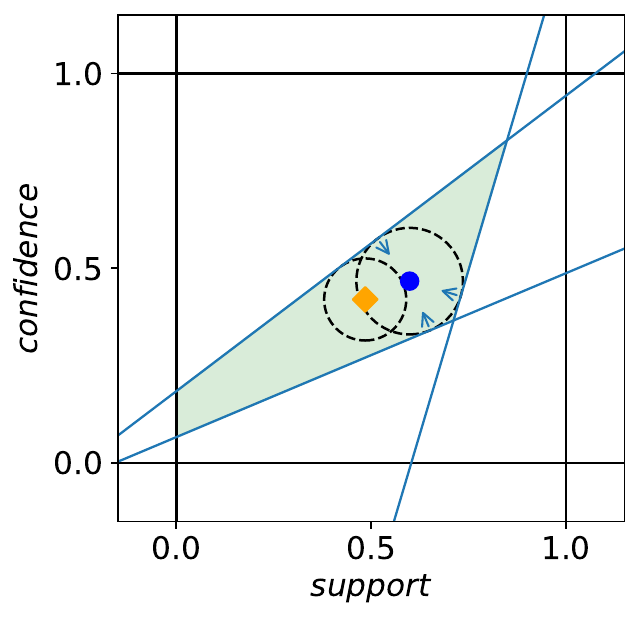}
    \caption{\(\versionSpace\) after \(3\) queries}
    \label{fig:vers_space}
  \end{subfigure}

  \begin{subfigure}[t]{\textwidth}
    \centering
    \includegraphics[width=.95\textwidth]{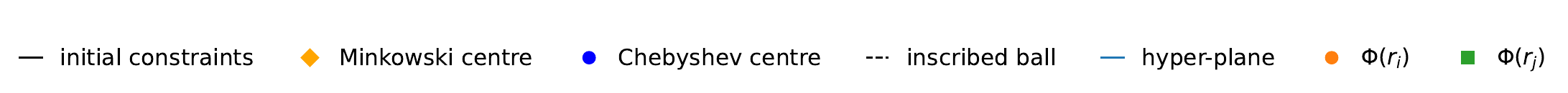}
  \end{subfigure}

  \label{fig:learning} 
\end{figure*}       

In the case of a \(2\)-additive Choquet integral, the utility function can be expressed as a \emph{linear separator} in an augmented feature space; the very same construction extends to any \(k\)-additivity. This formulation reveals useful geometric properties that are exploited both in learning and inference.
We define the augmented feature map:
\[
  \aug : \mathbb{R}^{d}\;\longrightarrow\;
         \mathbb{R}^{d+\binom{d}{2}},\qquad
  \aug(\feature)=
  \Bigl(
       \underbrace{\feature_{1},\dots,\feature_{d}}_{\text{singleton terms}},
       \,
       \underbrace{\min(\feature_{i},\feature_{j})}_{1\le i<j\le d}
  \Bigr),
\]
where the first $d$ coordinates represent the original interestingness measures, and each additional coordinate encodes the pairwise interaction via the minimum operator. This transformation captures all singleton and pairwise interactions, as required by the $2$-additive Choquet model.
Let \(\weights \in \mathbb{R}^{d + \binom{d}{2}}\) denote the vector of Möbius capacities:
\[
  \weights \;=\;
  \Bigl(
        \capacity(\{1\}),\dots,\capacity(\{d\}),\,
        \capacity(\{1,2\}),\dots,\capacity(\{d-1,d\})
  \Bigr)
\]
Then, evaluating the utility of a rule reduces to a dot product:
\[
  C_{\capacity}(\feature)
  \;=\;
  \bigl\langle \weights,\,\aug(\feature)\bigr\rangle
\]
Thus, learning a capacity consistent with user preferences is equivalent to identifying a point \(\weights\) within the feasible polytope \(\versionSpace\) in the space of dimension $d+\binom{d}{2}$, constrained by \eqref{eq:capacity-constraints} and all feedback constraints \eqref{eq:pref_constraints}. Rather than aiming to exactly recover the user's latent utility function$-$which may be noisy, ambiguous, or underdetermined from a limited number of comparisons$-$we approximate it with a representative utility function drawn from the current version space $\versionSpace$. 
To select a plausible model from this feasible region, we adopt geometric centers of $\versionSpace$ as surrogate models. The intuition is that a central model in $\versionSpace$ is more likely to generalize well across unseen queries, and can also guide the algorithm toward informative comparisons by identifying directions of greatest disagreement within the space.
However, computing the exact centroid of a polytope is \(\#\mathrm{P}\)-hard, we turn to two geometrically meaningful yet computationally efficient approximations:

\textit{- the \textbf{Chebyshev center}} is the center of the largest Euclidean ball fully contained within \(\versionSpace\). It can be efficiently computed by solving a small linear program~\cite{boyd2004_ac} (see Suppl. Mat.~\ref{sec:supp:cheby}). Geometrically, it identifies the point in $\versionSpace$ that maximizes the minimum distance to any of the polytope’s bounding hyperplanes. 

\textit{- the \textbf{Minkowski center}~\cite{den2024minkowski}} is the point \(x^\star \in \versionSpace\) that maximizes the symmetry of the feasible region. Formally, it is defined as the point that achieves the largest factor \(\lambda \in [0,1]\) such that for every \(p \in \versionSpace\), the reflected and scaled point \((1 - \lambda)x^\star + \lambda(2x^\star - p)\) remains within \(\versionSpace\). 
Unlike the Chebyshev center, which optimizes the distance to the boundary, the Minkowski center seeks for a globally averaged position rather than just maximizing local margin.

\noindent Our instantiation of Alg.~\ref{alg:gbs-choquet} is illustrated in Figures~\ref{fig:init_cons}--\ref{fig:vers_space}. 
We begin with the initial set of monotonicity and normalization constraints, from which we derive a first central point (Figure~\ref{fig:init_cons}). 
At this point, we compute the largest inscribed ball and then attempt to cut it (Figure~\ref{fig:max_ball}) with a hyperplane preference constraint. 
After identifying such a cutting constraint, we query the oracle and, based on its response, restrict the version space to the green region shown in Figure~\ref{fig:pref_cons}. 
After three queries, we obtain the version space depicted in Figure~\ref{fig:vers_space}, where the Minkowski and Chebyshev centers occupy distinct positions.

\section{Query Selection}
\begin{figure*}
    \includegraphics[width=\linewidth]{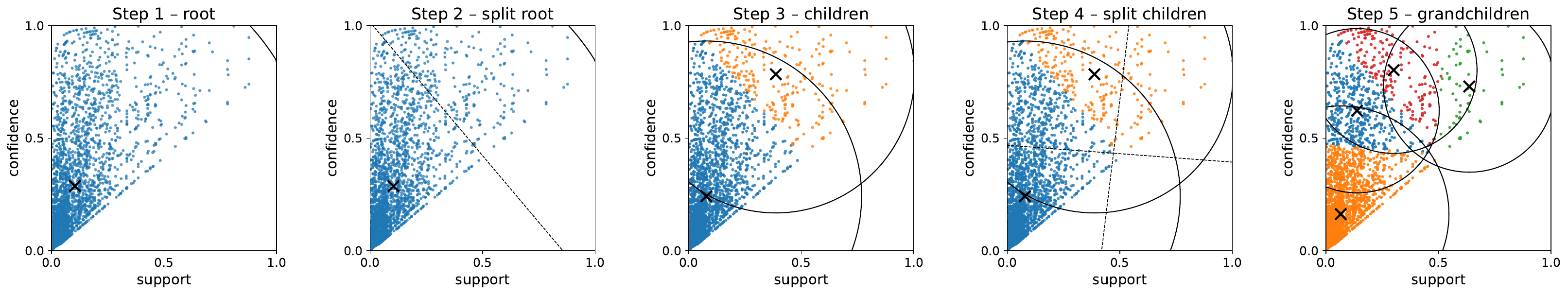}
    \caption{Ball-Tree Construction Algorithm}
    \label{fig:ball-tree-construct}
\end{figure*}

\begin{algorithm}[t] \footnotesize
\caption{\textsc{SelectQuery}$(\versionSpace, \rules) \rightarrow (\arule_i, \arule_j)$}
\label{alg:bestsplit}
\begin{algorithmic}[1]
\Require Version space $\versionSpace$, rule set $\rules$
\Ensure A rule pair $(\arule_i, \arule_j)$ yielding an interesting query or $\varnothing$
\State $c \gets \textsc{ComputeCenter}(\versionSpace)$
\State $\radius_{\max} \gets \textsc{InscribedRadius}(\versionSpace, c)$
\State $(\arule_i, \arule_j) \gets \textsc{SearchPair}(\rules, c, \frac{\radius_{\max}}{2})$
\State \textbf{if} $\dist(\arule_i, \arule_j, c) \le \radius_{\max}$ \textbf{then} \Return $(\arule_i, \arule_j)$
\State \textbf{else} \Return $\varnothing$
\end{algorithmic}
\end{algorithm}

\begin{algorithm}[t] \footnotesize
\caption{\textsc{SearchPair}$(\treeRoot,\,q,\,\tau)$}
\label{alg:bnb-pairwise}
\begin{algorithmic}[1]
\State $\heap\gets\emptyset, \quad$ \textsc{Push}$(\heap,\ \treeRoot.l,\ \treeRoot.r,\ q,\ \tau)$ \Comment{Ball-Node Min-Heap}
\State $(\arule_i^\star,\arule_j^\star)\gets\varnothing,\quad \dist^\star\gets+\infty$
\While{$\heap\neq\emptyset$}
    \State $(c_s,\LB,A,B,\UB)\gets$ \textsc{Pop}$(\heap)$ 
    \State \Return $(\arule_i^\star,\arule_j^\star)$ \textbf{if} $\dist^\star\le\tau$ \textbf{or} $\textsc{any}(A,B)$ \textbf{if} $\UB\le\tau$
    \If{$A,B$ are leaves}                             \Comment{\(O(n^2)\) exact check}
        \State $(p,s)\gets$ \textsc{ExactLeafEval}$(A,B,q)$
        \State \textbf{if} $s<\dist^\star$ \textbf{then} $(\arule_i^\star,\arule_j^\star,\dist^\star)\gets(p,s)$
    \Else                                                 \Comment{recursion}
    \ForAll{$a\in\{A.l, A.r\}$} \ForAll{$b\in\{B.l, B.r\}$}
        \If{$a\neq\varnothing$ \textbf{and} $b\neq\varnothing$}
            \State \textsc{Push}$(\heap,a,b,q,\min(\tau, \dist^\star))$      \Comment{cross-pair}
        \EndIf
    \EndFor\EndFor
    \If{$A.l\neq\varnothing$ \textbf{and} A.r$\neq\varnothing$}
        \State \textsc{Push}$(\heap, A.l, A.r, q, \min(\tau, \dist^\star))$   \Comment{intra-pair in $A$}
    \EndIf
    \If{$B.l\neq\varnothing$ \textbf{and} B.r$\neq\varnothing$}
        \State \textsc{Push}$(\heap, B.l, B.r, q, \min(\tau, \dist^\star))$   \Comment{intra-pair in $B$}
    \EndIf

    \EndIf
\EndWhile
\State \Return $(\arule_i^\star,\arule_j^\star)$
\end{algorithmic}
\end{algorithm}

\begin{algorithm}[t]
\footnotesize
\caption{\textsc{Push}$(\heap,\,a,\,b,\,q,\,\tau)$}
\label{alg:push}
\begin{algorithmic}[1]
\Procedure{Push}{$\heap, a, b, q, \tau$}
    \State $c_s \gets \dist_q\bigl(a.\text{center},\,b.\text{center}\bigr)$
          \Comment{distance of node centres to $q^\perp$ (Def.~\ref{def:pair-hyp-dist})}
    \State $\ell \gets \LB(a,b,q)$; \quad $u \gets \UB(a,b,q)$
    \If{$\ell > \tau$}                    \Comment{cannot beat the target}
        \State \Return
    \ElsIf{$u \le \tau$}                  \Comment{entire pair already good enough}
        \State \textsc{Insert}$\bigl(\heap,\;(0,\;\ell,\;a,\;b,\;u)\bigr)$  \Comment{sentinel key}
    \Else                                 \Comment{ordinary push}
        \State \textsc{Insert}$\bigl(\heap,\;(c_s,\;\ell,\;a,\;b,\;u)\bigr)$
    \EndIf
\EndProcedure
\end{algorithmic}
\end{algorithm}

Algorithm~\ref{alg:gbs-choquet} seeks a rule pair \((\arule_i, \arule_j) \in \rules^2\) such that the feature difference vector \(q = \Phi(\arule_i) - \Phi(\arule_j) \in \queries\) defines a hyperplane that bisects the version space \(\versionSpace\) into two polytopes of equal volume. As this perfect balance condition is hard to meet in practice, we relax it and instead search for interesting queries, defined as follows:
\begin{definition}
A query \((r_i, r_j) \in \rules\times \rules\) is said to be \emph{interesting} if the vector \(q = \Phi(r_i) - \Phi(r_j)\) defines a hyperplane \(q^\perp\) orthogonal to \( q \) that intersects the version space \(\versionSpace\), i.e.,
\[
    \exists\, \weights, \weights' \in \versionSpace \quad \text{such that} \quad \langle \weights, q \rangle \cdot \langle \weights', q \rangle < 0.
\]
\end{definition}

\noindent
\textit{Geometric intuition.}
The version space $\versionSpace$ is a convex polytope that contains all weight vectors $\weights \in \mathbb{R}^d$ consistent with the preference comparisons observed so far.
Each weight vector $\weights \in \versionSpace$ defines a utility function over rules, assigning a score 
\(
    \langle \weights, \Phi(r) \rangle 
\)
to every rule $r$ . 
Comparing two rules $r_i$ and $r_j$ under a given $\weights$ amounts to evaluating the sign of the product 
\(
    \langle \weights, q \rangle.
\)
If this sign varies across \( \weights \in \versionSpace \), then the models in \(\versionSpace\) are uncertain about the comparison between \( r_i \) and \( r_j \).
From a geometric perspective, the hyperplane \( q^\perp \) partitions the model space into two half-spaces: (1) one where \( r_i \) is preferred (i.e., \( \langle \weights, q \rangle > 0 \)), and (2) one where \( r_j \) is preferred (i.e., \( \langle \weights, q \rangle < 0 \)). If this hyperplane intersects the version space, then some \( \weights \in \versionSpace \) prefer \( r_i \), while others prefer \( r_j \), indicating a region of uncertainty. Algorithm~\ref{alg:bestsplit} presents a geometric method to efficiently find interesting queries. It starts by identifying a central point \(\centerModel \in \versionSpace\), located deep within the version space, and computes the largest radius \(\radius_{\max}\) of a ball centered at \(\centerModel\) that lies entirely within \(\versionSpace\). Any hyperplane \(q^\perp\) located within a distance less than \(\radius_{\max}\) from this center is guaranteed to intersect the version space, and therefore represents a query that reveals useful information, reducing model uncertainty.

\begin{definition}[Distance of a centre to a comparison hyperplane]\label{def:pair-hyp-dist}
Let \(r_i,r_j\in\rules\) be two rules, let \(\centerModel\in\versionSpace\) be the current central model, and set
\(
  q \;=\; \Phi(r_i)-\Phi(r_j).
\)
The \emph{distance} from \(\centerModel\) to the hyperplane \(q^{\perp}\) (i.e.\ the set of models that regard \(r_i\) and \(r_j\) as equally good) is defined as
\[
  \dist_{\centerModel}\!\bigl(\Phi(r_i),\Phi(r_j)\bigr)
  \;:=\;
  \frac{\bigl|\langle \centerModel,q\rangle\bigr|}{\|q\|_2}.
\]
\end{definition}

\noindent
Intuitively, \(q\) is the normal vector of the “indifference” hyperplane separating the two rules.  
If the center \(\centerModel\) lies \emph{close} to that hyperplane, it assigns nearly the same utility to \(\Phi(r_i)\) and \(\Phi(r_j)\); the preference between them is therefore uncertain.  
Conversely, a \emph{large} distance means the model strongly favors one rule over the other.  
The quantity \(\dist_{\centerModel}\) thus serves as a natural \emph{informativeness score}: pairs with small distance are the most ambiguous to \(\centerModel\) and, asking the user to choose between them is likely to split the version space into two sub-regions of roughly equal volume. Since computing this distance for all \(\mathcal{O}(|\rules|^2)\) rule pairs is infeasible, we propose an adapted version of the branch-and-bound (BnB) scheme of \citet{BallTree} that leverages geometric properties to estimate tight lower and upper bounds on this distance over groups of rules (represented as balls in feature space). These bounds enable branch-and-bound pruning: entire groups of pairs can be discarded early if their minimal distance exceeds the current best candidate's or the acceptance threshold \(\tau = \radius_{\max}\), thereby avoiding expensive evaluations and significantly accelerating the search.



\noindent
\textit{Ball-Tree Construction.} For the pruning to be efficient, the different balls in the ball tree need to be most disjoint. 
During construction we therefore split each parent ball by step 1 in Fig.~\ref{fig:ball-tree-construct} selecting two points that are far apart inside it, step 2 in Fig.~\ref{fig:ball-tree-construct} assigning every remaining point to the nearer pivot, and step 3 in Fig.~\ref{fig:ball-tree-construct} wrapping each resulting subset in its own minimal-radius ball.
This recursive “farthest‐point” partitioning drives sibling balls away from one another, producing branches that are nearly disjoint, step 5 in Fig.~\ref{fig:ball-tree-construct}.

\noindent
\textit{Tight Ball-Tree Bounds.} 
Given two feature-space balls \(\ball_1 = \ball(c_1, r_1)\) and \(\ball_2 = \ball(c_2, r_2) \subset\features \), and a reference direction vector \(\centerModel \in \versionSpace\), we define the total radius \(\rho = r_1 + r_2\) and displacement vector \(d = c_2 - c_1\). The following theorem provides tight geometric bounds on the pairwise distance to the hyperplane defined by \(\centerModel\), used for pruning in Algorithm~\ref{alg:bnb-pairwise}.

\begin{theorem}[Bounding Distance to a Query Hyperplane]\label{thm:ball-lower-corrected}
Let \(\gamma = \|\centerModel\|\) and \(\delta = \langle d, \centerModel\rangle\). Then for all \(\feature_1 \in \ball_1\) and \(\feature_2 \in \ball_2\) with \(\feature_1 \ne \feature_2\),
\[
   \UB(\ball_1,\ball_2,\centerModel)
   \;\;\ge\;\;
   \dist_{\centerModel}(\feature_1,\feature_2)
   \;\;\ge\;\;
   \LB(\ball_1,\ball_2,\centerModel),
\]
where
\[
   \LB(\ball_1,\ball_2,\centerModel) :=
   \begin{cases}
       0, & |\delta| \leq \rho\gamma \\[6pt]
       \gamma \cdot \frac{|\delta - \rho \gamma|}{\|d\gamma - \rho \centerModel\|}, & \text{otherwise}
   \end{cases}
\]
\[
   \UB(\ball_1,\ball_2,\centerModel) :=
   \begin{cases}
       \gamma \cdot \frac{|\delta - \rho \gamma|}{\|d\gamma - \rho \centerModel\|}, & \delta \leq 0 \\[6pt]
       \gamma \cdot \frac{|\delta + \rho \gamma|}{\|d\gamma + \rho \centerModel\|}, & \delta \geq 0
   \end{cases}
\]
\end{theorem}

\begin{proof}
We begin by taking the Minkowsky difference of \(\ball(\centerBall_1, \radius_1)\) with \(\ball(\centerBall_1, \radius_1)\) which results in \(\ball(d, \rho) = \{\centerBall_2 - \centerBall_1 + z \mid \|z\| \le \rho\}\). 
\\\noindent Regarding the \emph{lower bound} we can adapt the bound from~\cite{BallTree} to the un-normalized case of Def.\ref{def:pair-hyp-dist}. Let us distinguish between two scenarios, one in which the hyperplane defined by \(\centerModel^\perp\) intersects \(\ball(d, \rho)\) and one in which it does not. Intersection necessitates that the distance from the center \(d\) to the hyper-plane \(\centerModel^\perp\) be less than \(\rho\). If the intersection condition is met, there exists a point in \(\ball(d, \rho) \cap \centerModel^\perp\) rendering the distance null. When the hyper-plane and the ball are disjoint, we seek for the nearest point to the hyper-plane achievable in the ball. Such a point \(v = d + z, \|z\| \le \rho\) minimizes the parallel component of \(d\) w.r.t. \(\centerModel\). This point is attained at \(z^\star = -\rho \frac{\centerModel}{\gamma}\) with \(\gamma = \|\centerModel\|\). Evaluating the distance from \(\centerModel\) to \(v^\perp\), we get:
\begin{equation*}
    \frac{|\langle d-\rho \frac{\centerModel}{\gamma}, \centerModel\rangle|}{\|d - \rho\frac{\centerModel}{\gamma}\|} = \gamma\frac{|\langle d, \centerModel\rangle - \rho \gamma|}{\|d\gamma - \rho \centerModel\|}
\end{equation*}
We derive \(|\langle d, \centerModel\rangle| \leq \rho\gamma\) from the intersection condition. When searching for an \emph{upper bound} we seek the furthest achievable point from the hyper-plane \(\centerModel^\perp\). Yet again we distinguish between two scenarios: one where the center \(d\) lies in the positive half-space \(\{\langle x, \centerModel\rangle \geq 0, x \in \features\}\) and one where it lies in the negative half-space. Contrary to the lower bound, we seek a point \(v = d + z, \|z\| \le \rho\) that maximizes the parallel component of \(d\) w.r.t. \(\centerModel\). This point is attained at \(z_+^\star = \rho \frac{\centerModel}{\gamma}\) with \(\gamma = \|\centerModel\|\) in the positive half-space scenario and at \(z^\star = -\rho \frac{\centerModel}{\gamma}\) in the negative half-space scenario. We obtain the desired upper bound by evaluating the distance from \(\centerModel\) to \(v^\perp\).
\end{proof}
\noindent
\textit{Algorithm~\ref{alg:bnb-pairwise}.}
The procedure follows a best-first search strategy guided by \(c_s\): the minimum distance from two node centroids
to the query hyperplane \(q^\perp\).
Algorithm~\ref{alg:push} maintains the visitation order.
Given the target threshold \(\tau=\min\{\dist^\star,r_{\max}\}\),
it (i) discards every node pair whose lower bound already exceeds
\(\tau\) and (ii) immediately reinserts any pair whose \emph{upper}
bound is below \(\tau\), using a sentinel key so it will be popped
next.
The search stops when the heap is empty or when the current best
distance \(\dist^\star\le\tau\).
When the most promising entry \((A,B)\) is popped at iteration \(i\),
two cases arise.
If both nodes are leaves, the algorithm evaluates the pair-to-hyperplane
distance for every rule combination in \(A\times B\) (Def.~\ref{def:pair-hyp-dist})
and updates \((\arule_i^\star,\arule_j^\star)\) whenever a smaller
distance is found.
Otherwise it expands the branch by pushing to the heap the four
cross-child pairs \((A_l,B_l),(A_l,B_r),(A_r,B_l),(A_r,B_r)\) as well as
the two intra-child pairs \((A_l,A_r)\) and \((B_l,B_r)\).

\section{Experiments}
For all experiments we employ the following statistical features: Yule’s Q, Cosine, Kruskal’s \(\tau\), Added Value, and the Certainty Factor (see Suppl. Mat.~\ref{sec:exp_prot} for more experimental configuration details).

\noindent
\textbf{Statistical‐\textbf{rule–measure} oracle (\(\mathbf{\phi}\)).}
Tan et al.~\cite{tan_selecting_2004} showed that many statistical interestingness measures fall into a few \emph{equivalence classes}: measures in the same class rank rules in exactly the same order, whereas measures from \emph{different} classes do not. The measure \(\phi\) belongs to one such class.  
For our feature set we deliberately picked five measures, each drawn from a class \emph{different} from \(\phi\)’s.  
Hence every single feature is, by construction, \emph{agnostic} to \(\phi\).  
Their \emph{aggregate}, however, need not be: a non-linear aggregator such as the Choquet integral could exploit interactions between the features and recover information that none of them carries alone.  
By evaluating Alg.~\ref{alg:gbs-choquet} against the \(\phi\) oracle we therefore test whether the Choquet integral can “bridge the gap” from individually \(\phi\)-agnostic features to a collective representation that aligns with the \(\phi\) ranking.

\noindent
\textbf{\textbf{Surprise} against a maximum--entropy oracle.} 
If all you know about a \emph{transactional database} are the single–item frequencies $p_i$, the most neutral assumption is that items occur \emph{independently}.  
The probability that every item in a rule $X\!\Rightarrow\!Y$ appears together is $\prod_{i\in X\cup Y}p_i$, so in $n$ transactions we \emph{expect}
\(
f_{\text{exp}} = n\prod_{i\in X\cup Y}p_i
\)
such co-occurrences.  
The oracle compares this to the \emph{observed} count $f_{\text{obs}}$ and assigns the score
\(
S = \log_2\frac{f_{\text{obs}}}{f_{\text{exp}}} = \log_2 f_{\text{obs}} - \log_2 f_{\text{exp}}.
\)
By evaluating against this surprise oracle we test the adaptivity of Alg.~\ref{alg:gbs-choquet} to a highly agnostic setting. 

\begin{figure*}[htbp]
  \centering

  {
    \captionsetup[subfigure]{aboveskip=2pt, belowskip=0pt}

    \begin{subfigure}[b]{\textwidth}
      \centering
      \includegraphics[width=\textwidth]{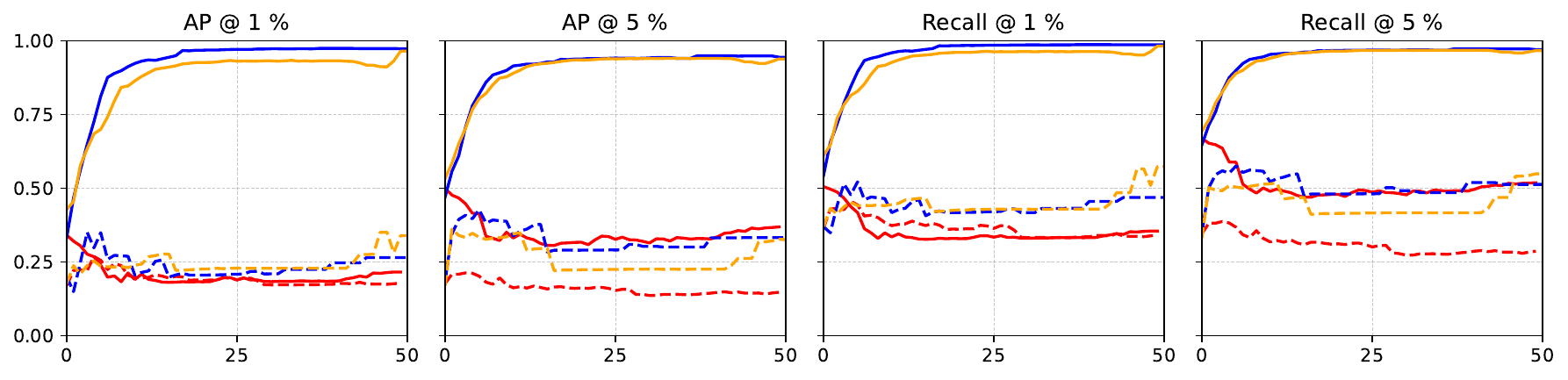}
      \caption{\(1\)-additive Choquet integral — colour = method, line style = oracle.}
      \label{fig:choq-1}
    \end{subfigure}

    \begin{subfigure}[b]{\textwidth}
      \centering
      \includegraphics[width=\textwidth]{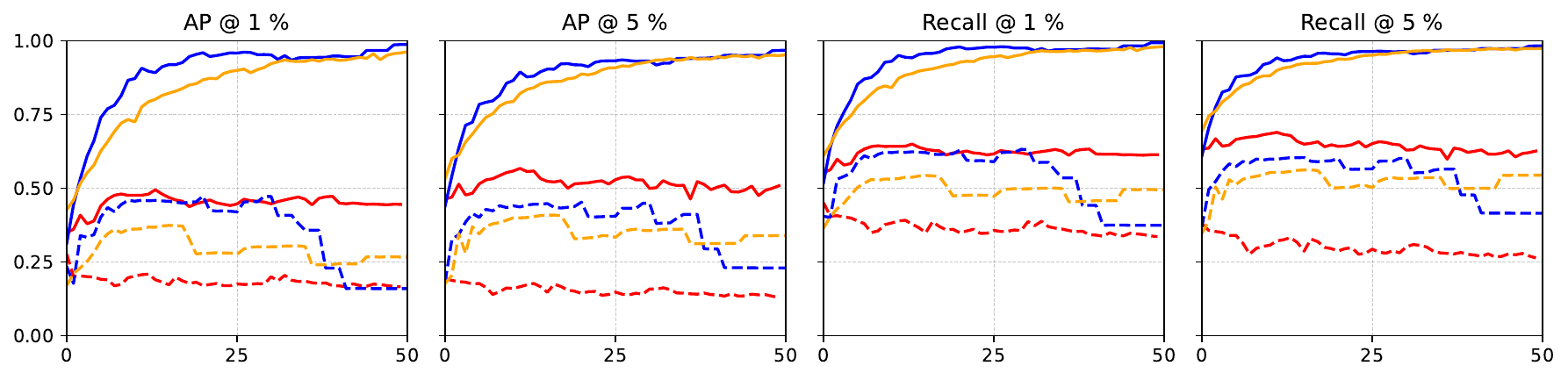}
      \caption{\(2\)-additive Choquet integral.}
      \label{fig:choq-2}
    \end{subfigure}

    \begin{subfigure}[b]{\textwidth}
      \centering
      \includegraphics[width=\textwidth]{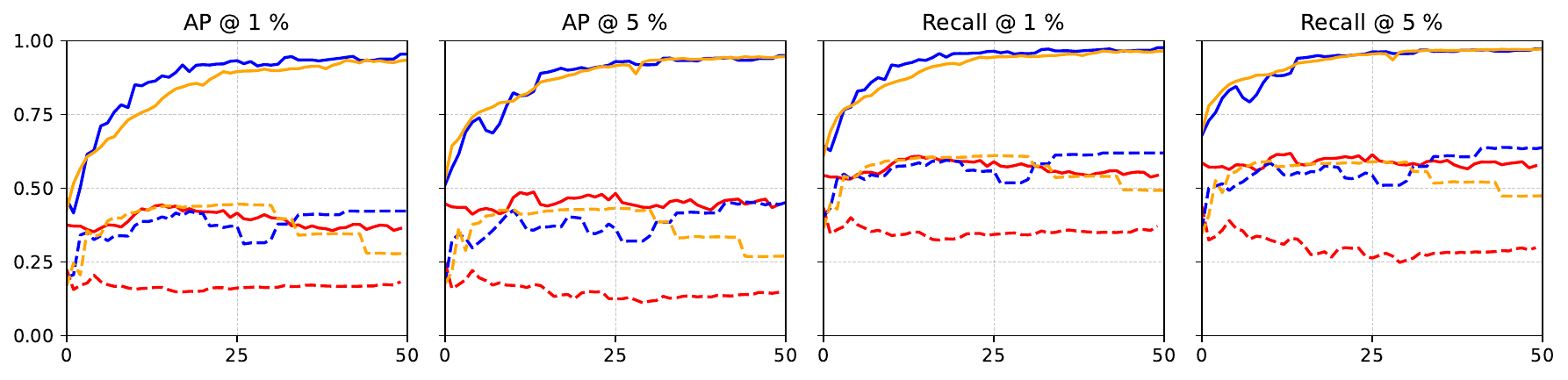}
      \caption{\(3\)-additive Choquet integral.}
      \label{fig:choq-3}
    \end{subfigure}

    \begin{subfigure}[b]{\textwidth}
      \centering
      \includegraphics[width=\textwidth]{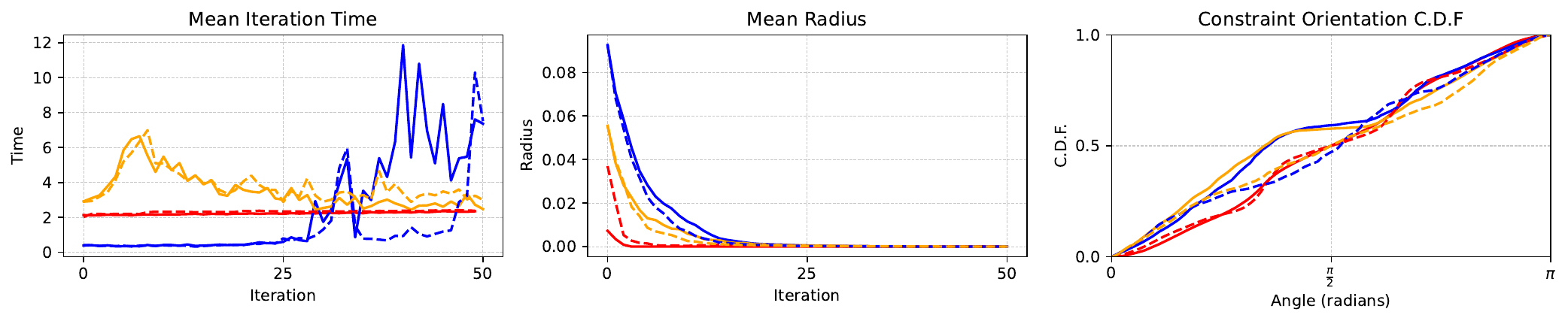}
      \caption{Qualitative and version-space metrics for the \(2\)-additive Choquet integral.}
      \label{fig:stats-2}
    \end{subfigure}
  }

  \makebox[\textwidth][c]{%
    \begin{subfigure}[b]{0.30\textwidth}
      \includegraphics[width=\linewidth]{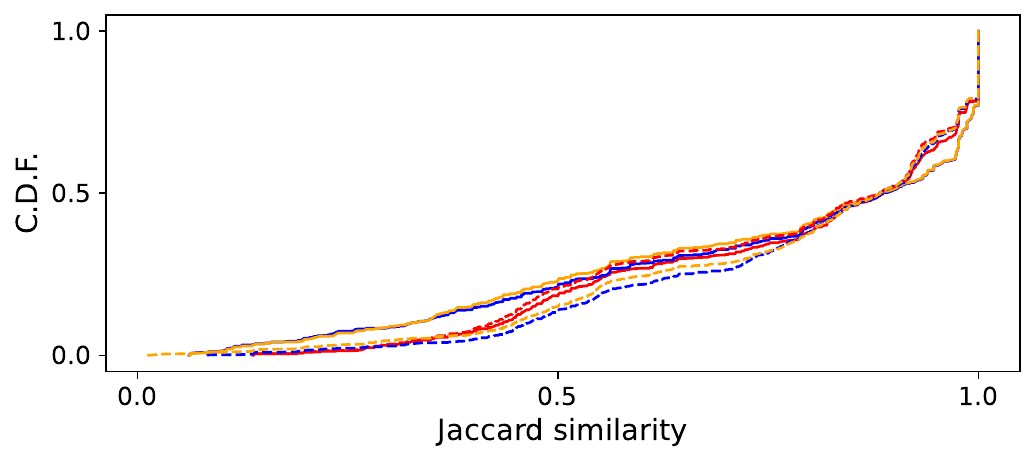}
      \caption{\(1\)-additive}
      \label{fig:jaccard-1}
    \end{subfigure}\hfill
    \begin{subfigure}[b]{0.30\textwidth}
      \includegraphics[width=\linewidth]{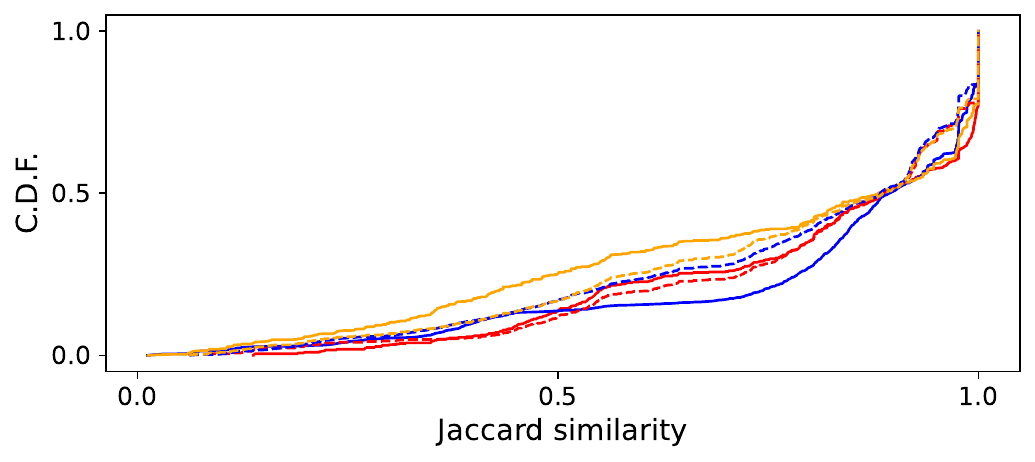}
      \caption{\(2\)-additive}
      \label{fig:jaccard-2}
    \end{subfigure}\hfill
    \begin{subfigure}[b]{0.30\textwidth}
      \includegraphics[width=\linewidth]{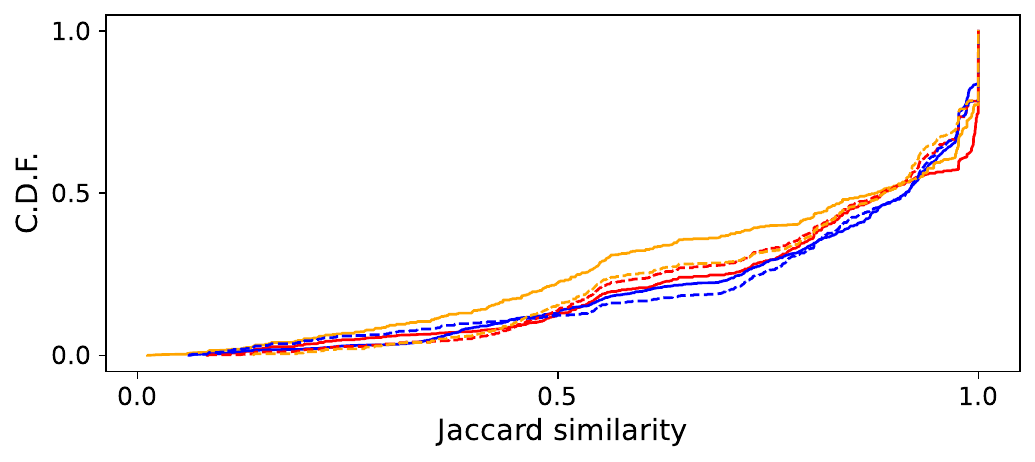}
      \caption{\(3\)-additive}
      \label{fig:jaccard-3}
    \end{subfigure}%
  }

  \vspace{0.6em}
  \makebox[\textwidth][c]{%
    \includegraphics[width=0.8\textwidth]{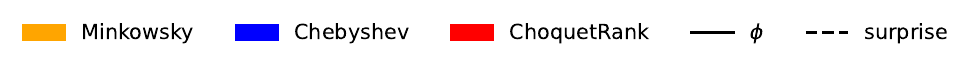}%
  }

  \caption{Choquet-integral performance and diversity analyses across additive orders.}
  \label{fig:choquet-full}
\end{figure*}

\noindent
\textbf{Consequences of increasing additivity.} 
Figures~\ref{fig:choq-1}–\ref{fig:choq-3} reveal a clear trade-off between model expressiveness and convergence speed. 
In the 1-additive setting, the Minkowski and Chebyshev centres (orange and blue solid lines) converge fastest and attain the highest scores on every metric for the \(\phi\) oracle, showing that even a linear aggregation of \(\phi\)-agnostic features can closely approximate \(\phi\)’s ranking. 
Moving to the 2- and 3-additive settings (Fig.~\ref{fig:choq-2} and~\ref{fig:choq-3}) slows convergence but raises performance on the highly agnostic surprise oracle (dashed lines). 
Hence, greater additivity yields more expressive—and ultimately more accurate—models, but at the cost of slower convergence.

\noindent
\textbf{Query selection and iteration time.} 
Figure~\ref{fig:stats-2} reports the mean wall-clock time per iteration (averaged over folds and datasets) for the 2-additive experiments. 
As \CR\ chooses queries by random sampling, its running time is essentially constant, which explains the near-flat curve. 
The branch-and-bound strategy behaves differently: as learning proceeds the radius of the largest admissible ball shrinks exponentially, and finding a rule pair whose separating hyperplane still intersects that ball becomes increasingly costly, thus the search time grows. 
Inexplicably, the Minkowski center shows a rise–fall pattern which may be due to its evolving position in the version space, starting near a boundary and gradually settling inward. To verify that the algorithm does not keep cutting the version space with nearly parallel hyperplanes, we also plot the cumulative distribution of pairwise angles between successive preference constraints. 
The almost-uniform CDF in Figure~\ref{fig:stats-2} indicates that the cutting directions remain well diversified throughout the run (Cf. Supp.~\ref{sec:exp_prot}).

\noindent
\textbf{Top-\(k\) diversity and overall performance} (Cf. Supp.~\ref{sec:exp_prot})
Figures~\ref{fig:jaccard-1}–\ref{fig:jaccard-3} plot the Jaccard similarity (lower is better) between the transaction covers of the top-15 rules.
All methods follow the same general pattern, yet a clear trend towards \emph{greater diversity} appears as we move from 1- to 3-additive models.
Thus higher additivity not only improves accuracy but also yields rule lists with more diversified covers—another manifestation of the expressiveness-versus-convergence trade-off. 

\noindent
\textbf{Implementation details.}
The entire pipeline is implemented in \textsc{Python} and released under an open-source license here\footnote{\url{https://github.com/Tudor1415/KDD-UMC-Geometric-AL}}. 
All raw datasets together with their evaluations can be downloaded here\footnote{\url{https://drive.google.com/drive/folders/132GJjjRn1ypJYsFCil8GY51zZHWUU8Ji}}.

\section{Conclusion}
Across more than \(1\,000\) query iterations per oracle–method pair, our geometry-driven strategy consistently outperforms the heuristic baseline of \CR~\cite{ChoquetRank} on precision, recall, and convergence rate at low cut-offs.
The branch-and-bound strategy entails higher per-iteration cost, but it deterministically finds a query hyperplane that intersects the current version space, whereas the random sampling in \CR\ offers no such guarantee. The additional computation therefore yields consistently better rankings and more informative queries—a favorable trade-off in most scenarios.

\smallskip
\noindent
\textbf{Acknowledgments}
We gratefully acknowledge Laurent Truffet for the insightful discussions on polytopal geometry. %
This research project was financially supported by the French national research project FIDD under grant agreement ANR-24-CE23-0711.

\clearpage
\bibliographystyle{ACM-Reference-Format}
\bibliography{main}

\clearpage
\section{Supplementary Material}
\subsection{Experimental Details}
\label{sec:exp_prot}
\paragraph{Experimental Protocol} We carried out our experiments on the five UCI datasets listed in Table~\ref{tab:dataset-stats}. 
For every dataset we mined, with \emph{Choco--Mining}~\cite{ChocoMining}, up to ten million \emph{minimal, non-redundant} association rules that satisfy a confidence of at least \(99\%\) and a support of at least \(10\). 
Each rule was evaluated on five statistical interestingness measures: \textbf{Yule’s \(Q\)}, \textbf{Cosine similarity}, \textbf{Goodman--Kruskal’s \(\tau\)}, \textbf{Added Value}, and the \textbf{Certainty Factor}. 
Rules with identical five-dimensional feature vectors were deduplicated, but dominated rules were \emph{not} removed so as not to bias the study against oracles that might legitimately prefer a dominated rule over a Pareto-optimal one. 
After de-duplication we capped the rule set of every dataset at \(100\,000\) and evaluated all methods under \emph{three-fold cross-validation}. 
For both ChoquetRank and our branch-and-bound learner, the loop stopped as soon as the oracle replied 'indifferent' or \(0\), since neither algorithm currently supports indifference feedback.
\paragraph{Jaccard Top-k similarity plots} To measure how diverse the highest–ranked rules are, we proceeded as follows.  
First, every rule was ranked with the final iteration Choquet capacity for its oracle–center–fold setting.  
Next, we kept the \emph{top fifteen} rules and turned each one into a binary cover vector that flags the transactions it captures.  
For every pair of covers $(\mathbf v_i,\mathbf v_j)$ we then computed the Jaccard similarity  
\[
J(\mathbf v_i,\mathbf v_j)=
\frac{\lVert \mathbf v_i \land \mathbf v_j\rVert_1}
     {\lVert \mathbf v_i \lor \mathbf v_j\rVert_1},
\]
where $\land$ and $\lor$ are element-wise logical \textsc{and} / \textsc{or}.  
The empirical distribution of these $J$ values—aggregated over folds—underlies the similarity curves shown in the paper.
\paragraph{Constraint orientation plot} For every query we record the \emph{augmented} difference vector
\(\mathbf h=\aug(\Phi(r_i))-\aug(\Phi(r_j))\)
To derive a linear constraint in the \(d-1\)-dimensional where the version space polytope is fully dimensional (does not contain any equality constraints), we substitute the weight sum equality, yielding  
\[
\mathbf a
  \;=\;
  (h_1,\dots,h_d)-h_{d+1}\mathbf 1,
  \qquad
  b
  \;=\;
  -h_{d+1},
\]
such that the preference reads \(\mathbf a^{\top}\mathbf w \le b\).
Normalizing \(\mathbf a\) gives the unit normal
\(\mathbf n=\mathbf a/\lVert\mathbf a\rVert_2\).  
Collecting all such normals for a given \textit{(oracle, centre)} pair,
forming every unordered pair \((\mathbf n_i,\mathbf n_j)\), and computing
\(\theta_{ij}=\arccos(\mathbf n_i^{\top}\mathbf n_j)\) produces a set of
angles whose empirical C.D.F.\ is the constraint–orientation curve in Figure~\ref{fig:stats-2}.

\begin{table}[htbp]
\centering
\caption{Statistics of the datasets used in the experiments.}
\begin{tabular}{lrrr}
\toprule
Dataset & $|\items|$ & $|\SDB|$ & $|\rules|$\\
\midrule
mushroom  & 95  &   8\,124 & 100\,000 \\
tictactoe & 28  &     958 &  23\,473 \\
magic     & 80  &  19\,020 & 100\,000 \\
credit    & 73  &  30\,000 & 100\,000 \\
twitter   & 214 &  49\,999 & 100\,000 \\
\bottomrule
\end{tabular}
\label{tab:dataset-stats}
\end{table}

\subsection{The Minkowski Center}
\label{sec:supp:mink}
It maximizes a measure \(\lambda\) of central symmetry and can be computed using the following LPs from~\cite{den2024minkowski}.
\begin{figure}[htbp]
\centering
\begin{minipage}[t]{0.46\linewidth}
  \centering
  \textbf{Minkowski–centre LP}\\[2pt]
  \begin{align}
  \max_{w,\lambda}\;&\lambda\\
  \text{s.t.}\quad
      &Aw=(1+\lambda)\,b \\[2pt]
      &Cw-\lambda\,\delta\le d \\[2pt]
      &\lambda\ge 0
  \end{align}
\end{minipage}%
\hfill
\begin{minipage}[t]{0.46\linewidth}
  \centering
  \textbf{Facet–minimum LPs for }$\boldsymbol{\delta}$\\[12pt]
  \begin{align}
  \delta_i=\;
  \min_{y}\;&c_i^{\top}y\\
  \text{s.t.}\quad
      &Ay=b\\[2pt]
      &Cy\le d
  \end{align}
\end{minipage}
\caption{LPs used to compute the Minkowski center and the facet offsets $\delta$.}
\label{fig:minkowski-lps}
\end{figure}

\noindent
The program in Fig.~\ref{fig:minkowski-lps} maximizes the symmetry factor $\lambda$ of the polytope $C=\{x\mid Ax=b,\;Cx\le d\}$. 
Constraint $Aw=(1+\lambda)b$ ensures that the scaled point $w/(1+\lambda)$ still satisfies all equalities. 
The inequality $Cw-\lambda\delta\le d$ tightens every facet by the amount $\lambda\delta$, guaranteeing that every $\lambda$-scaled reflection of $C$ about the candidate center remains inside $C$. 
The bound $\lambda\ge 0$ rules out negative symmetry radii. 
At optimality, $\lambda^\star$ equals the Minkowski symmetry and $x^\star=w^\star/(1+\lambda^\star)$ is a Minkowski center.

\subsection{The Chebyshev Center} 
\label{sec:supp:cheby}
It maximizes the radius \(r\) of the largest inscribed ball in a polytope.
\begin{figure}[htbp]
\centering
\textbf{Chebyshev–centre LP}\\[2pt]
\begin{align}
\max_{x,r}\;& r \\[4pt]
\text{s.t.}\quad
      &Ax=b \\[2pt]
      &Cx + r\,\kappa \;\le\; d \\[2pt]
      &r \ge 0
\end{align}
\caption{LP used to compute the Chebyshev center.}
\label{fig:chebyshev-lp}
\end{figure}
\noindent
In Fig.~\ref{fig:chebyshev-lp} $x\in\mathbb R^{n}$ is the candidate center of the largest inscribed Euclidean ball and $r\in\mathbb R_{+}$ its radius.  
The equalities $Ax=b$ keep the center on the affine subspace common to every feasible point.  
For each facet $c_i^{\top}x\le d_i$, the term $\kappa_i=\lVert c_i\rVert_{2}$ measures how far that facet shifts when the ball grows; the constraint $Cx+r\kappa\le d$ therefore pulls every facet inward by exactly $r$ so that the entire ball stays inside the polytope.  
Maximizing $r$ then yields the Chebyshev radius $r^{\star}$ and its associated center $x^{\star}$.

\end{document}